\documentclass[conference]{IEEEtran}
\IEEEoverridecommandlockouts
\usepackage{cite}
\usepackage{amsmath,amssymb,amsfonts}
\usepackage{algorithmic}
\usepackage{graphicx}
\usepackage{textcomp}
\usepackage{xcolor}
\usepackage{fancyhdr,lipsum}
\def\BibTeX{{\rm B\kern-.05em{\sc i\kern-.025em b}\kern-.08em
    T\kern-.1667em\lower.7ex\hbox{E}\kern-.125emX}}
    
\newcommand{\MYfooter}{\smash{
\hfil\parbox[t][\height][t]{\textwidth}{\centering
\thepage}\hfil\hbox{}}}

\makeatletter

\def\ps@IEEEtitlepagestyle{%
\def\@oddhead{\parbox[t][\height][t]{\textwidth}{\centering
2026 IEEE International Conference on Contemporary Computing and Communications (InC4)\\
}\hfil\hbox{}}%
\def\@evenhead{\scriptsize\thepage \hfil \leftmark\mbox{}}%
\def\@oddfoot{979-8-3195-1915-3/26/\$31.00 \textcopyright 2026 IEEE \hfil
\leftmark\mbox{}}%
\def\@evenfoot{\MYfooter}}

\begin{document}
% paper title sairam sairam sairam sairam sairam sairam
%\title{Machine-Style Handwriting Generation with Diffusion}

%%sairam sairam sairam sairam sairam saram sairam sairam sairam sairam sairam sairam sairam sairam sairam sairam sairam sairam sairam sairam sairam sairam sairam sairam sairam sairam
%%\title{NoiseFormer - Noise Diffused Symmetric Attention Transformer}

\title{NoiseFormer -- Noise Diffused Symmetric Attention Transformer*\thanks{This work has been submitted to the IEEE for possible publication. Copyright may be transferred without notice, after which this version may no longer be accessible.}}

\author{
     \IEEEauthorblockN{Phani Kumar, Nyshadham}
     \IEEEauthorblockA{\textit{Intel Corporation} \\
          phani.kumar.nyshadham@intel.com} \and
     \IEEEauthorblockN{Jyothendra Varma, Polisetty V R K}
     \IEEEauthorblockA{\textit{Intel Corporation} \\
     polisetty.v.r.k.jyothendra.varma@intel.com}
}

\maketitle

%% sairam sairam sairam sairam sairam sairam sairam sairam sairam sairam sairam sairam sairam sairam sairam sairam
\begin{abstract}
      Transformer architecture has been very successful long runner in the field of Deep Learning ($DL$) and Large Language Models ($LLM$) because of its powerful attention-based learning and parallel-natured architecture. As the models grow gigantic in terms of memory footprint, difficulties in fitting the model on a device like a $GPU$ or an AI accelerator give rise to the need for multiple computing devices thereby escalating the computing cost. This increased training/inference cost paved the way for optimizing the memory footprint of the models in terms of activation memory as well as model size reduction. In this paper, we start analyzing one of the Attention architectures aimed at model-size reduction called Symmetric Dot-Product Attention \cite{sym_attention} (referred to as Symmetric Attention) and propose a novel unified model architecture called Noise Diffused Symmetric Attention Transformer to enhance the model's performance. While maintaining the memory gains of Symmetric Attention, with minute overhead in terms of model parameters and compute in the form of noise diffused attention, the proposed model recovers the loss in performance due to Symmetric Attention compared to conventional attention as depicted through the inference-time sampling and GLUE scores on downstream tasks.

\end{abstract}

\section{Introduction} %%sairam sairam sairam sairam sairam sairam
Attention-based transformers are finding their applications from moderately complex language tasks like translation, summarization, classification to complex text generation tasks like chain-of-thought reasoning, coding, long context understanding. The cost of training scales with the complexity of the problem in terms of model size and compute budget needed to model the short-term and long-term contextual relationships within the input. Typically, with attention mechanism, the high dimensional input features are compressed into latent vectors and stacking multiple attention blocks model highly complex non-linear probability functions using hierarchical learning (for example, from words to phrases to sentences).

More specifically, the decoder-only Transformers like GPT-models have become the heart of agentic AI systems which are on the rise. The ease of parallelization of attention-based transformers across powerful computing devices like GPUs and AI accelerators is one of the main reasons for them to thrive. Despite their success, the exponential growth in the parameters (from 10's of billions to 100's of billions and towards trillions) and context lengths of Transformer-based models have necessitated advanced architectural techniques to reduce the memory footprint and computational overhead. Broadly, these optimization strategies can be categorized into sparse-attention mechanisms, model memory reduction techniques, or a hybrid of both. Examples like Longformer \cite{longformer2020} and Reformer \cite{reformer2020} focus on reducing the memory footprint of attention and activation matrices by deploying Sparse Attention techniques like sliding-window attention and locality-sensitive-hashing. On the other hand, model memory reduction techniques such as Grouped-Query Attention \cite{gqa2023} and Multi-Query Attention \cite{mqa2021} drastically compress the parametric memory footprint by sharing key and value projection heads across multiple attention queries. Finally, SOTA architectures like DeepSeek-V2 \cite{deepseek_mla2024} integrates both these approaches by utilizing a sparse Mixture-of-Experts (MoE) framework to reduce active compute parameters besides advanced latent attention mechanisms for substantial activation memory (KV cache) compression.

There has been extensive ongoing research around Multi-Head-Attention $MHA$ mechanism to reduce the model capacity in terms of parameters unveiling newer attention architectures. In this paper, we start off with a model-size reduction technique around Multi-Head-Attention $MHA$ mechanism called Symmetric Dot-Product Attention \cite{sym_attention} using Query ($Q$) and Key ($K$) as same and progress towards using generative AI techniques thus improving the performance.

In this paper, we proceed to model generic attention as noisy version of symmetric attention to improve the model's representative capacity. The rest of the paper is organized as follows. Section II talks about Symmetric Dot-Product Attention. Section III talks about the proposed idea of generalizing attention as noisy version of Symmetric Attention. Section IV discusses the experiments performed with proposed model architecture. Section V presents objective performance results of the proposed model as compared to the Symmetric Attention mechanism along with the base model also, during inference. Section VI talks about the future work with concluding remarks.

\section{Symmetric Dot-Product Attention - Sparse Attention} %%sairam sairam sairam sairam sairam sairam
In Transformer architectures, the core of scaled dot-product attention ($SDPA$) lies in the three matrices - Query ($Q$), Key ($K$) and Value ($V$). The cross-correlation between $Q$ and $K$ produces the raw attention scores which are normalized and used for weighting the $V$ vectors to generate latent representation of input tokens \cite{sym_attention}. % as depicted in Figure (1).

%\begin{figure}[t]
%     \begin{center}
%          \centerline{\includegraphics[width=0.25\columnwidth]{sdpa_sai.jpg}}
%          \caption{Conventional Attention in Transformers \cite{sym_attention}}
%     \end{center}
%     \vspace{-2em}
%\end{figure}

The mathematical formulation of $SDPA$ as a function of $Q, K, V$ projections of input token embeddings, as part of Multi-Head-Attention (MHA) depicted in Figure (1), is defined as

\begin{equation}
Attn(Q, K, V) = softmax(\frac{QK^T}{\sqrt{d_k}})V
\end{equation}

To improve the training efficiency of Transformer-based models and to reduce their resource consumption, \cite{sym_attention} has proposed to use an alternative compatibility function $A$ to compute the compatibility between two tokens $x$ and $y$. The scaled dot-product attention $QK^T$ can be viewed in terms of compatibility function between the $Q$ and $K$ projections of $x$ and $y$ respectively as follows:

\begin{equation}
A(x,y) = Q(x)K(y)^T
\end{equation}

By enforcing the feature sharing between $Q(x)$ and $K(y)$ for input tokens $x$ and $y$, \cite{sym_attention} proposed to make $Q=K$, thus reducing the memory footprint of the model significantly. This results in compatibility function being symmetric as defined below:

\begin{equation}
A_{sym}(x,y) = Q(x)Q(y)^T
\end{equation}

While the above symmetric compatibility function enables feature sharing between $Q$ and $K$ projections, the model suffers from lack of expressiveness for asymmetric feature relationships amongst input tokens. This results in the loss of model performance reflected in the results enumerated in \cite{sym_attention}. To improve the performance, \cite{sym_attention} has also proposed pair-wise compatibility function which adds extra memory overhead to the Symmetric Attention mechanism.

As mentioned in the last section, we deviate from this point in our paper where we apply concepts of generative AI to enhance the performance of Symmetric Dot-Product Attention with near-zero model-size and compute overhead. The next section elaborates on the proposed idea. It is to be noted that the proposed model architecture is discussed and elaborated in the context of decoder-only GPT-like auto-regressive models, and it is to be understood that the proposed model can be applied to any transformer model (encoder/decoder/encoder-decoder) built on $SDPA$ attention blocks.

\section{Proposed Model Architecture} %%sairam sairam sairam sairam sairam sairam sairam sairam
As discussed in the last section, Symmetric Dot-Product Attention (referred to as Symmetric Attention, for brevity) is a special case of Generic Attention where $Q = K$. If we look deeper as to how generic attention is computed, referring to Figure (1), we define 2 matrices $W_q$ and $W_k$ which are the projection matrices from input embedding space $X$ to the Query and Key spaces $Q$ and $K$ respectively such that $W_q, W_k : R^{hidden} \rightarrow R^{head}$ which compute $Q$ and $K$ as below:

\begin{align}
Q &= X*W_q \nonumber \\
K &= X*W_k  \nonumber \\
QK^T &= XW_q*W_k^TX^T = XW_{qk}X^T
\end{align}

where $W_{qk} = W_qW_k^T$. 

\begin{figure}[t]
     \begin{center}
          \centerline{\includegraphics[width=0.7\columnwidth]{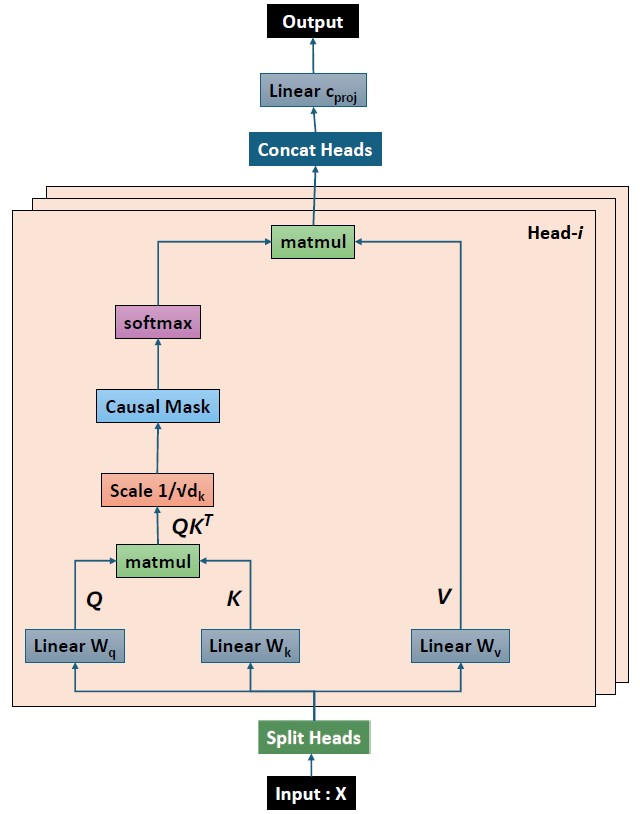}}
          \caption{Conventional Multi-Head Attention ($MHA$) in Transformers}
     \end{center}
     \vspace{-2em}
\end{figure}

For symmetric attention \cite{sym_attention}, we make $W_q = W_k$ as depicted in Figure (2) and hence the $SDPA$ computes $QQ^T$ which becomes symmetric and is computed as follows:

\begin{align}
Q &= X*W_q \nonumber \\
K &= X*W_q = Q \qquad\text{(since } W_q = W_k\text{)} \nonumber \\
QK^{T} &= Q Q^{T} = X W_q W_q^{T} X^{T} = XW_{qq}X^T
\end{align}

where $W_{qq} = W_qW_q^T$.

\begin{figure}[t]
     \begin{center}
          \centerline{\includegraphics[width=0.7\columnwidth]{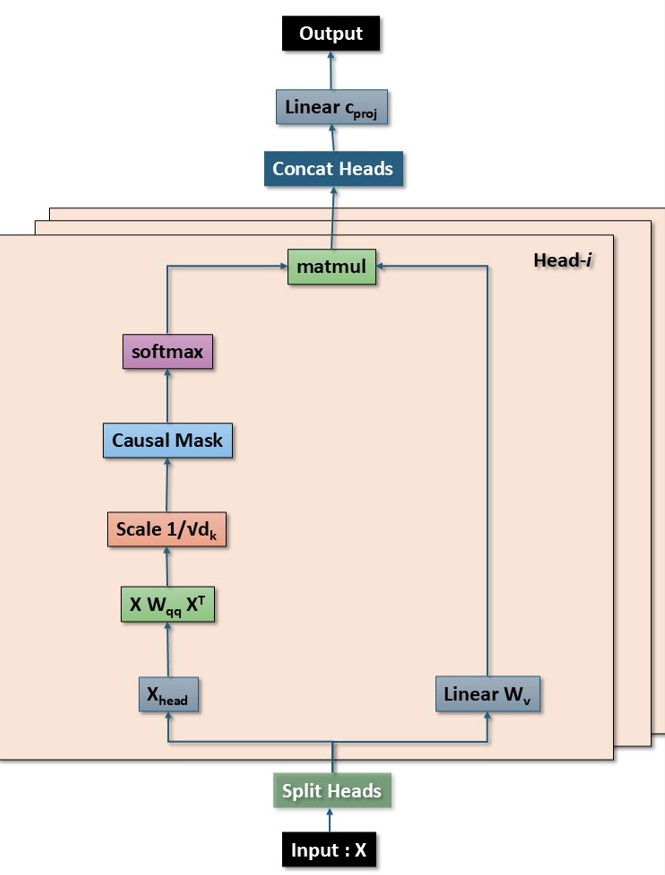}}
          \caption{Symmetric Multi-Head Attention \cite{sym_attention}}
     \end{center}
     \vspace{-2em}
\end{figure}

\subsection{Generic Attention as Noisy Symmetric Attention} %%sairam sairam sairam sairam sairam
The generic attention mechanism has different $Q$ and $K$ projections computed by $W_q$ and $W_k$ trainable matrices, which are further deployed for computing normalized attention scores using $softmax$ as in Equation (1). Intuitively, symmetric attention suffers from lack of expressiveness because of enforcing the symmetric constraint as $W_q = W_k$ leading to $Q=K$. To bridge the performance gap manifested in terms of the representative power of the model, the key idea in this paper is to model generic attention as a noisy version of the symmetric attention as below:

\begin{equation}
QK^T = QQ^T + noise^{\prime}
\end{equation}

We proceed to approximate the generic attention with the symmetric attention added with noise ($noise^{\prime}$) and parameterize the noise across the whole training set. We propose a kind of unsupervised learning set-up for optimal parameterization of additive noise. As an example, the statistics of additive noise can be efficiently estimated using criteria like Maximum Likelihood ($MLE$). After sufficient training, we can approximate generic attention scores as noised symmetric attention scores with the optimal noise statistics achieved. The noise modelling can vary from simple Gaussian to complex manifolds for the application at hand. In this paper, we have done a coarse modelling of the additive noise ($noise^{\prime}$) being single mode Gaussian $\mathcal{N}(-\mu, \sigma^2)$. Mathematically, any noise distribution can be modeled as Gaussian mixture with several mixture components and hence single-mode Gaussian is selected as a basic-level approximation for the purpose of analysis and the concept proposed is no longer limited to single-mode Gaussian distribution of noise.

\subsection{Generic Attention score Estimator} %%sairam sairam sairam
Once the noise estimates are obtained, the next task is to arrive at a good estimator of generic attention scores from symmetric attention scores. For doing this, we start with the additive noise model of generic attention as in Equation (6) and reframe the problem as follows:

\begin{align}
QK^T &= QQ^T + noise^{\prime} \nonumber \\
\implies QQ^T &= QK^T - noise^{\prime}
\end{align}

From the above equations, we have done reverse modelling of symmetric attention scores being noised version of generic attention scores. The noise being subtracted ($noise^{\prime}$) from the generic attention scores can be considered as an additive noise ($noise$) for the reverse modelling, whose mean is negative of mean of $noise^{\prime}$ and whose variance is the same as the estimated $noise^{\prime}$ in the previous section. Which means $noise$ is modelled as single mode Gaussian $\mathcal{N}(\mu, \sigma^2)$. The reframed equation is as follows:

\begin{align}
QQ^T &= QK^T + noise
\end{align}

With the above modelling, we can come up with an estimator for generic attention scores $QK^T$ as a function of symmetric attention scores $QQ^T$. Assuming $QK^T$ is also a random variable, we proceed to extract a good estimate of the generic attention score from the symmetric attention score. As an example, if we wish to estimate the generic scores in Minimum-Mean-Square-Error ($MMSE$) sense, the estimator becomes conditional mean estimator as described below:

\begin{align}
\widehat{QK^T}_{MMSE} &= E\{QK^T|QQ^T\}
\end{align}

where $E\{X|Y\}$ is the conditional mean of $X|Y$. As another example, if we wish to get a Maximum A Posteriori Probability ($MAP$) estimate, the estimator will be as follows:

\begin{align}
\widehat{QK^T}_{MAP} &= arg \hspace{0.1cm} max_{QK^T}{f(QK^T|QQ^T)}
\end{align}

where $f(x|y)$ is the conditional probability density of $X$ given $Y$. Computing the conditional density $f(QK^T|QQ^T)$ is challenging and can only be approximated across a variety of probability density functions of $QK^T$. The choice of cost function to be used for estimating $QK^T$ from $QQ^T$ has to be evaluated with respect to the final objective of auto-regressive modelling in the case of decoder-only models. This means, it must be an iterative process for achieving the final objective of decoder-only models, which is not really convenient. To make the process more scalable and for training efficiency, we propose a unified model architecture to improve the performance of symmetric attention-based transformers \cite{sym_attention} elaborated in the next section.

\subsection{Noise Diffused Symmetric Attention Transformer (NoiseFormer) - Proposed Unified Model architecture} %%sairam sairam sairam sairam sairam sairam sairam
As described in the last sections, when we proposed generic attention scores as noisy version of the symmetric attention scores, we are performing two steps - \textit{optimal noise modelling} and \textit{optimal generic attention score estimator}. To this end, as a first step, we need to perform unsupervised learning for noise modelling. As the second step, we have to choose an optimal estimator for a variety of cost functions to achieve the final objective of auto-regressive models. What if, the actual objective for estimator is such that it does not yield a closed form solution for generic attention score estimator. Also with complex noise modelling and density function approximation of $QK^T$, the estimator design can become more challenging to derive.

To address the challenges described above, we propose a unified model architecture called Noise Diffused Symmetric Attention Transformers (also called as NoiseFormer) where we tackle both the noise modelling and generic attention score estimation together at the same time. The choice of estimator is also addressed in an efficient manner with the proposed model architecture. Noise Diffused Symmetric Transformer builds upon the Symmetric Attention Transformer where noise is injected into the symmetric attention block and gets added to $QQ^T$. The added noise can range from simple uni-modal Gaussian to complex multi-modal Gaussian Mixture and is completely the choice of the model designer and application at hand. The idea is to enable the model to learn the underlying empirical noise density across the data samples by approximating it as multi-modal Gaussian mixture density. As an example, in this paper, we explain the concept by modelling the noise modelling as a simple uni-modal Gaussian distribution $\mathcal{N}(\mu, \sigma^2)$. To estimate reliably the generic attention scores from symmetric attention scores, instead of iteratively estimating across a variety of cost functions, we learn the optimal cost function also within the model itself such that the final loss of auto-regressive modelling is minimized. Figure (3) depicts the Noise diffused Symmetric Attention head where noise ($\mathcal{N}(\mu, \sigma^2)$) is added after scaled symmetric attention computation $QQ^T$ followed by Softmax to produce probability scores which are further used in matrix multiplication with Value ($V_{out}$) matrix. In the foregoing discussion, it is to be noted that Softmax follows the noise addition, for every head, though not explicitly stated.

\begin{figure}[t]
     \begin{center}
          \centerline{\includegraphics[width=0.6\columnwidth]{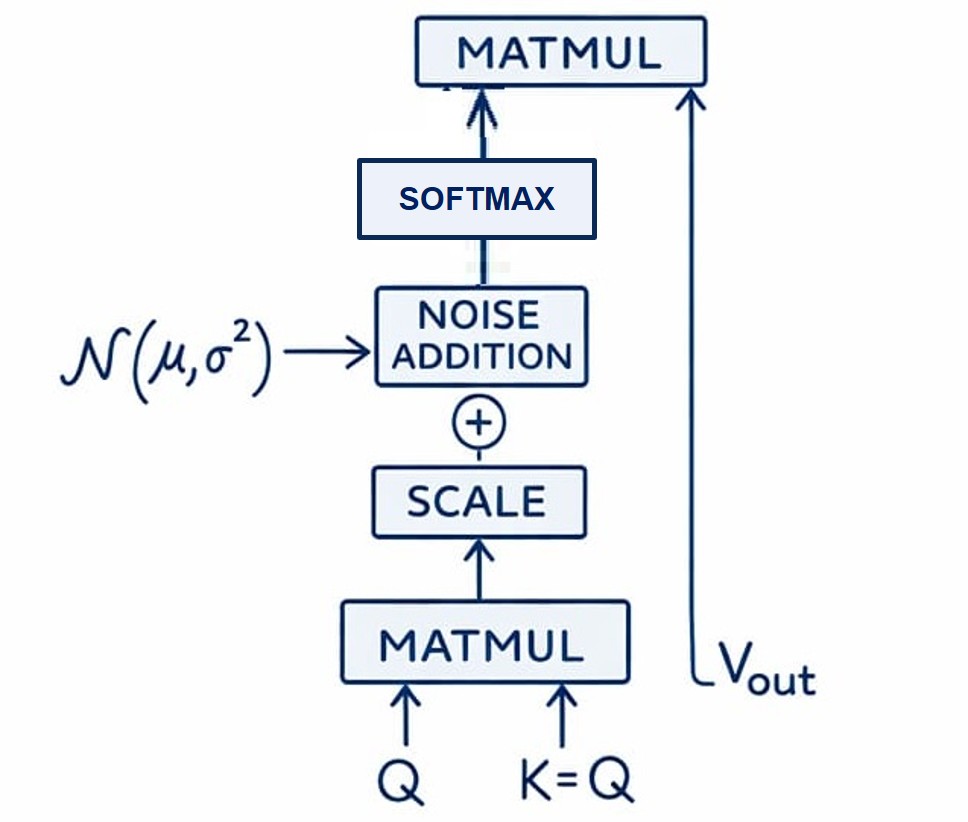}}
          \caption{Single head Attention in Proposed NoiseFormer}
     \end{center}
     \vspace{-2em}
\end{figure}

\begin{figure}[ht]
     \begin{center}
          \centerline{\includegraphics[width=0.7\columnwidth]{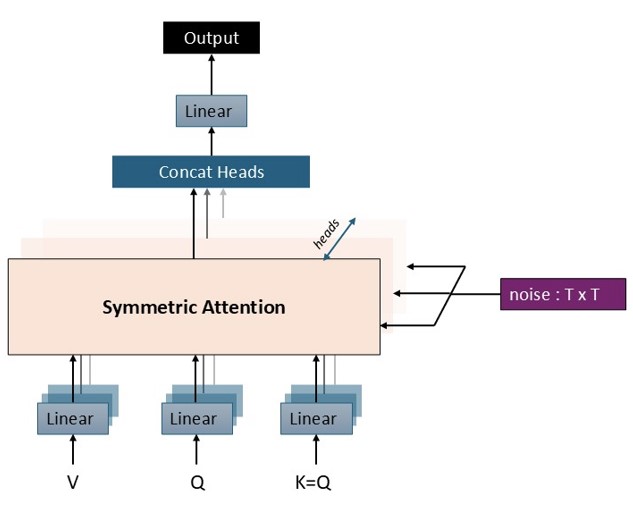}}
          \caption{All-heads Shared Noisy Symmetric Attention (All-heads SNSA) in a NoiseFormer layer}
     \end{center}
     \vspace{-2em}
\end{figure}

Figure (4) depicts a variant of the proposed NoiseFormer where same noise tensor is added across the attention heads in a Multi-Head-Attention ($MHA$) setting for a transformer layer. The mean and variance of noise added to all the heads change for every transformer layer as shown in Figure (5). From the figure, it is seen that a noise tensor of size $T \times T$ is generated with $\mathcal{N}(\mu_{i}, \sigma_{i}^2)$ gets added to each head's $QQ^T$ in the $i^{th}$ transformer layer in the model. The same noise tensor sampled from $\mathcal{N}(\mu_{i}, \sigma_{i}^2)$ is added to all the $j$ heads inside the $MHA$ block of the $i^{th}$ transformer layer as described in the following equation.

\begin{align}
     Attn_{noisy_{shared}} &= (Q_i^{(j)})(Q_i^{(j)})^T + \mathcal{N}(\mu_i, \sigma_i^2)
\end{align}

\begin{figure}[t]
     \begin{center}
          \centerline{\includegraphics[width=0.7\columnwidth]{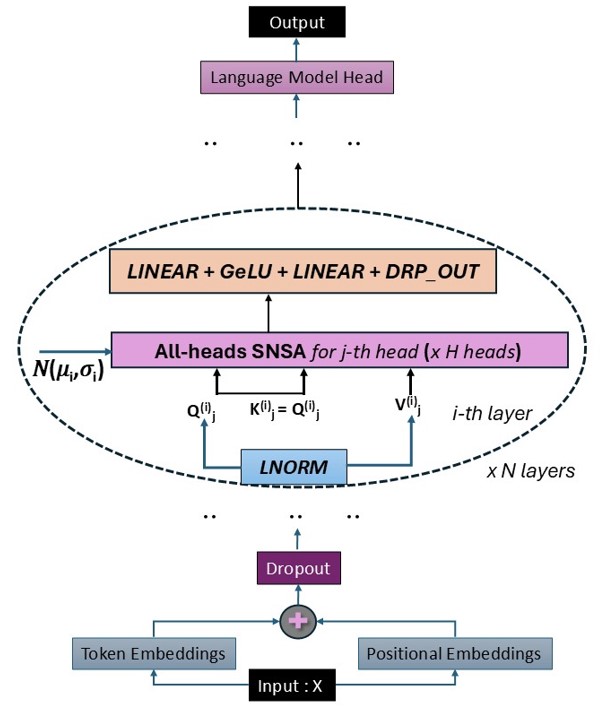}}
          \caption{All-heads SNSA across layers in a NoiseFormer model}
     \end{center}
     \vspace{-2em}
\end{figure}

Figure (6) depicts a deeper variation of the NoiseFormer where every head in a $MHA$ block of a transformer layer is added with noise of a different mean and variance. Figure (7) shows noise tensor of size $N \times T \times T$ is generated with $\mathcal{N}(\vec{\mu}_{i}, \vec{\sigma}_{i}^2)$ gets added to the $MHA$ block of $i^{th}$ transformer layer where $\vec{\mu}_{i}$ and $\vec{\sigma}_{i}^2$ are a $1 \times N$ vectors each, where $N$ is the number of heads inside $MHA$ block. In this case, for $j^{th}$ attention head of $MHA$ block inside $i^{th}$ transformer layer, the noisy attention scores are computed as 

\begin{align}
Attn_{noisy_{per-head}} &= (Q_i^{(j)})(Q_i^{(j)})^T + \mathcal{N}(\mu_i^{(j)}, (\sigma_i^{(j)})^2)
\end{align}

\begin{figure}[ht]
     \begin{center}
          \centerline{\includegraphics[width=0.7\columnwidth]{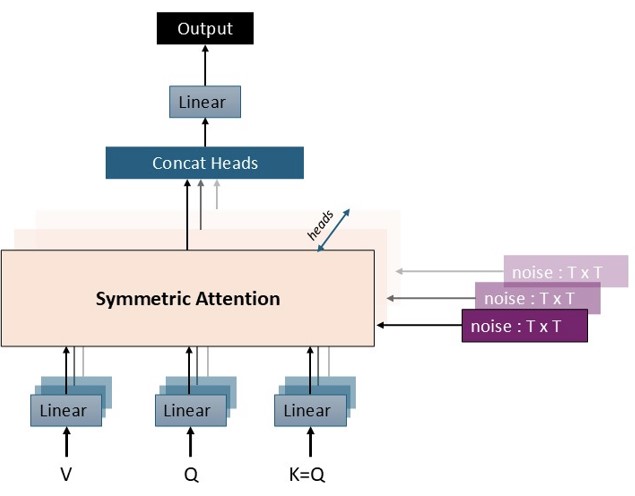}}
          \caption{Per-head Shared Noisy Symmetric Attention (Per-head SNSA) in a NoiseFormer layer}
     \end{center}
     \vspace{-2em}
\end{figure}

\begin{figure}[t]
     \begin{center}
          \centerline{\includegraphics[width=0.7\columnwidth]{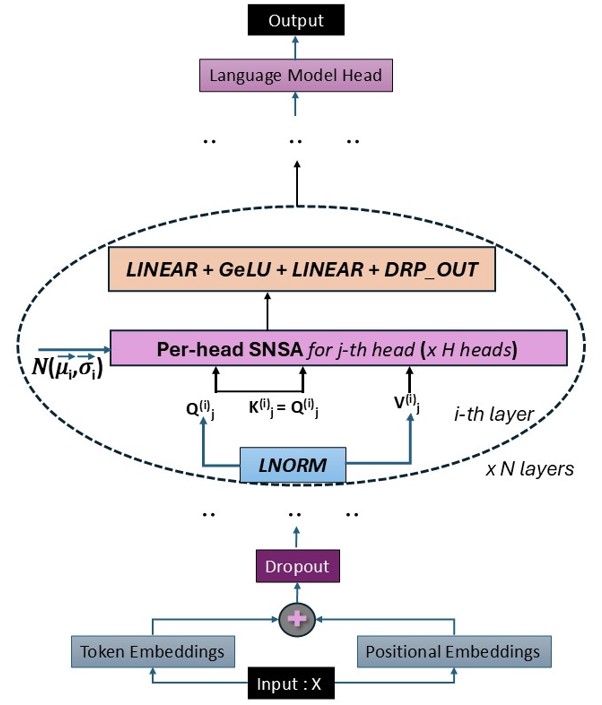}}
          \caption{Per-head SNSA across layers in a NoiseFormer model}
     \end{center}
     \vspace{-2em}
\end{figure}

The above explanation shows an example of a simple noise model and a couple of variants of noise injection into the $MHA$ blocks of NoiseFormer layers. It is to be noted that the proposed model is not limited to the noise model examples and the discussed variants.

\subsection{NoiseFormer - Global Loss function} %%sairam sairam sairam sairam
With the proposed NoiseFormer model, as described above, the aspects of noise modelling and generic attention score estimation are folded into the same proposed model. The objective function for the auto-regressive model is used as the same loss function for training the proposed model. The negative log-probability ($NLL$) for predicting a sequence of tokens $y_{1:T}$ parameterized by the model parameters ($\theta$) is given as

\begin{align}
\mathcal{L}_{\text{seq}}(y_{1:T}; \theta) = -\sum_{t=1}^T \log \hat{p}_\theta\big(y_t \mid y_{<t}\big) \\
\mathcal{L}_{\text{seq}}(y_{1:T}; \theta) = -\sum_{t=1}^T \Big( z_{t,y_t}(\theta) - \log\!\sum_{v\in\mathcal{V}} e^{z_{t,v}(\theta)} \Big)
\end{align}

where $\hat{p}_\theta\big(y_t \mid y_{<t}\big)$ is the model's probability of predicting $y_t$ at time  $t$ given the previous token sequence $y_{<t}$ and is given by the last softmax layer of the model in terms of logits $z_{t,v}$ for $v^{th}$ index of vocabulary $\mathcal{V}$ as 

\begin{align}
\hat{p}_\theta(y_t \mid y_{<t}) = \frac{\exp(z_{t,y_t}(\theta))}{\sum_{v\in\mathcal{V}} \exp(z_{t,v}(\theta))}
\end{align}   

The above loss function can also be written as a consequence of softmax-cross-entropy loss. Following mathematical theory of VAE models for generative AI, we constrain the noise statistics to not be very scattered in which case the output predicted sequence can be very off for the drawn noise samples. To enforce this constraint, we followed VAE encoder loss which aims to keep the distribution of the added noise as close as possible to standard uni-modal Gaussian distribution defined as $\mathcal{N}(0, 1)$. The KL divergence term with respect to standard Gaussian for a given Gaussian distribution $\mathcal{N}(\vec{\mu}, \vec{\sigma}^2)$ is given by

\begin{align}
KL_{loss} = 0.5\sum_k{{\mu_k}^2 + {\sigma}_{k}^2 - log{\sigma}_{k}^2 - 1}
\end{align}  

The $KL_{loss_{global}}$ can be computed as a summation of individual $KL_{loss}$ computed for every noise distribution across heads and across layers. The $KL_{loss_{global}}$ is weighted by a factor $\alpha$ and then added to the cross-entropy loss to compute the global loss ($loss_{global}$) as defined as:

\begin{align}
loss_{global} = \mathcal{L}_{\text{seq}}(y_{1:T}; \theta) + \alpha*KL_{loss_{global}}
\end{align}  
The value of $\alpha$ signifies the importance of constraint imposed to force the noisy distribution to be standard Gaussian. By increasing $\alpha$ the constraint is more enforced limiting the diversity of noise added and its bias. Too small $\alpha$ over-diversifies and increases the bias of the noise making confused inference at times. An optimal value of $\alpha$ is required to balance between the above extremes. 

Modelling the noise densities to be closer to standard Gaussian may appear over-restrictive in the sense of modelling the generic attention as noised symmetric attention, but for the sake of proving the concept, it is done likewise and the results are quite encouraging. Based on applications, desired multi-modal probability density mixtures can be used in the place of standard Gaussian and accordingly KL divergence loss is to be formulated. The value of $\alpha$ is also determined by the applications at hand and the desired accuracy. By training the Noiseformer to minimize the above $loss_{global}$ along with the noise parameters (mean vectors and variance vectors), the model learns to approximate the generic attention as noisy version of symmetric attention and also learns the optimal estimator of generic attention scores given the symmetric attention scores, thereby minimizing Negative log-likelihood of the auto-regressive model.

\subsection{Novel learning aspects} %%sairam sairam
With the proposed unified model, the $loss_{global}$ is constructed to include the reliable noise modelling and optimal generic attention estimation. Also, this formulation brings new learning aspects which are observed during inference. As the model learns to inject optimal noise to the Multi-Head-Attention heads of every transformer layer, the noise statistics learnt capture the global context of the data generated like mood and style of the data. Which means, during inference, given the noise vector, the model generates the tokens with some global context learnt from the added noise. The diversity of the samples generated is rich as compared to the conventional transformers because of the randomness of the noise being added during inference. In a nutshell, for any mood or style implicit to the noise sample generated at inference time, the model tries to adapt and provide meaningful completion.

%The completions for a few prompts sampled 3 times for the pre-trained models (Base GPT2 model \cite{nanogpt2023} pre-trained for $300B$ tokens and the proposed NoiseFormer with same Noise tensor added across all the heads within a $MHA$ block per transformer layer pre-trained for $\sim 49.152$ Billion tokens) are attached as part of Appendix. The temperature setting for inference is set as $0.8$ for random sampling of next token. For the 3 inference samples, for each prompt across the models, the maximum number of generated tokens is set as $100$.

As another variation of Noiseformer, we also propose to have a tiny FFN layer working on the input token sequence to encode the information into the latent vectors which are used to generate the additive noise, which is injected into the MHA attention blocks, thus learning input-conditioned noise statistics instead of independent statistics. This variation is different from Free transformers \cite{free} which hinges on latent driven generation as a different concept altogether.

\subsection{Model size - Memory and compute overhead} %%sairam sairam sairam
Let's consider a GPT-like decoder-only model with $L$ transformer layers, each layer with $N$ heads comprising a $MHA$ block. Let the input sequence be of length $T$ tokens and $H$ be the hidden dimension of the input embeddings which makes the input matrix to $MHA$ as $T \times H$. The projection matrices for generating Query ($Q$) and Key ($K$) matrices from input are $W_q$ and $W_k$ respectively. The sizes of $Q$ and $K$ matrices across all the heads is given by $T \times H$ where $Q$ per head and $K$ per head are computed to be $T \times \frac{H}{N}$ each. Using symmetric attention making $W_q = W_k$, the saving in parametric memory per transformer layer in $MHA$ block is $T \times H$. For $L$ layers, the memory saving is $L*T*H$ parameters.

\begin{table}[htbp]
\caption{Inference Time Memory and Compute Overhead comparison for $MHA$ blocks in GPT-like models for uni-modal Gaussian Noise-Diffused Attention} %%sairam sairam sairam
\begin{center}
\begin{tabular}{|p{3.0cm}|p{2.75cm}|p{2.0cm}|}
\hline
\textbf{Model} & \textbf{$Q$ and $K$ params for $MHA$ blocks} & \textbf{$QK^T$ for $MHA$ blocks} \\
\hline
$NanoGPT_{Base}$ & $O(2*T*H*L)$ & $O(T^2*H*L)$ \\
\hline
$NanoGPT_{SymAttn}$ & $O(T*H*L)$ & $O(T^2*H*L)$ \\
\hline
$NanoGPT_{Proposed}$ & $O(T*H*L + 2*N*L)$ & $O(T^2*H*L)$ \\
\hline
\end{tabular}
\end{center}
\end{table}

With the proposed NoiseFormer, the model memory overhead with respect to the Symmetric Transformers is just the mean and variance vectors for all the layers which is near-zero.

\subsection{Compute Kernel optimizations}
The compute overhead is majorly contributed by the generation and addition of noise matrix $T \times T$ per layer. During inference time, as an optimization, the noise matrices are generated a priori on the host with the frozen means and variances, and the noise tensors can be transferred onto the device during forward propagation. Also, to avoid extra addition of noise matrices with attention scores, the output of Matrix Multiplier Engine ($MME$) can be initialized with noise matrix as partial sum so that the matrix multiplier result accumulates onto the noise matrix. In the case of training, the above optimization of $MME$ can be applied in which case, a standard Normal noise tensor can be generated on the host a priori and transferred onto the device and multiplication with $\sigma$ and addition with $\mu$ can be done efficiently using BLAS-GEMM functions on the device. 

Table I enumerates the memory required for model parameters and inference-time compute overhead involved in the proposed model architecture using $GPT2$ as the base model. From the table, the model memory overhead for the proposed model with respect to symmetric attention is $O(2*N*L)$ for uni-modal uncorrelated Gaussian Noise diffusion and $O(2*N*L*M)$ for $M$-modal uncorrelated Gaussian Noise diffusion which is very insignificant while the compute overhead is nearly zero.

\subsection{Applications and Versatility}
The proposed NoiseFormer architecture can be applied to any Transformer-based models because it deals with the basic foundational attention mechanism of Transformer architecture. Not only to Language based models, but the proposed model concepts can also be applied to Vision Transformer models and Diffusion models which run on Attention as the core module. The proposed model architecture can be experimented with the Latent Multi Head Attention mechanism proposed in Deep-Seek model too.

\section{Experiments}
In this section, we elaborate on the experiments conducted on different model architectures for performance evaluation. As part of experimentation, we have pre-trained several models followed by fine-tuning guided for downstream tasks. Firstly, we pre-train models to measure validation loss during training to track model progress. We evaluate each model variant on GLUE benchmark to find model's accuracy on various downstream tasks.

The base model chosen for experimentation is $NanoGPT$ model \cite{nanogpt2023} from GPT2 family having 12 transformer blocks, each block consisting of 12 heads in multi head attention ($MHA$) with Layer-Normalization and GeLU activation. The embedding dimension for this model is chosen as $768$. The model size in terms of parameters for the $NanoGPT$ model is $\sim 123.5$ Million. 

On top of $NanoGPT$ base model, we applied techniques of Symmetric Attention based Transformers \cite{sym_attention} and call that model as Symmetric Attention model. Similarly we applied the proposed NoiseFormer architecture on top of $NanoGPT$ base model and obtained two variants of our proposed model for experimentation - NoiseFormer variant with same noise matrix added to all heads' raw attention scores inside $MHA$ block per layer (referred to as $V_1$) and the variant with noise matrix added per head to the raw attention scores inside $MHA$ block per layer (referred to as $V_2$). All the models are trained under the same conditions as elaborated further.

\subsection{Pre-training Dataset - Setup}
Openwebtext dataset is used for pre-training which has $8,013,769$ documents and approximately $300$ Billion tokens. Pre-training dataset is tokenized using GPT2 tokenizer. OpenWebText dataset is primarily English and is an open-source replication of the WebText dataset from OpenAI, that was used to train GPT2.

%The authors started by extracting all Reddit post urls from the Reddit submissions dataset. These links were de-duplicated, filtered to exclude non-html content, and then shuffled randomly. The links were then distributed to several machines in parallel for download, and all web pages were extracted using the newspaper python package. Using Facebook FastText, non-English web pages were filtered out.
%Subsequently, near-duplicate documents were identified using local-sensitivity hashing (LSH). Documents were hashed into sets of 5-grams and all documents that had a similarity threshold of greater than 0.5 were removed. The the remaining documents were tokenized, and documents with fewer than 128 tokens were removed. This left 38GB of text data (40GB using SI units) from 8,013,769 documents.

We have deployed 8 Intel Gaudi2 accelerators in a distributed training setup for pre-training the models. Training recipe used by us is sequence length of $1024$, gradient accumulation steps of $40$ with batch size $12$ which means each iteration uses $\sim 491,520$ tokens per back-propagation step. Weight-decay is $0.1$, iterations-per-evaluation are $200$ with evaluation interval as $1000$. Noise initialization of mean and standard deviation is from normal distribution with noise factor as $0.01$. AdamW optimizer is deployed with learning rate as $0.0006$, $beta_1=0.9$, $beta_2 = 0.95$, $grad_{clip}$ at $1.0$, learning-rate ($lr$) decay enabled with warm-up iterations as $2000$ with minimum learning rate as $0.00006$. Learning-rate decay scheduler (cosine with warm-up) uses decay-iterations as $600K$. Training is done in FP32 precision. Cross-entropy loss with KL Divergence (weight factor is $0.000005$) is used as loss function and validation loss is tracked for model progress. Our models are trained for $100K$ steps while GPT2 based $NanoGPT$ model is pre-trained for $300K$ steps.

\subsection{Fine-tuning for downstream tasks}
Pretrained models are fine-tuned on GLUE dataset for benchmarking to find models' Natural Language Understanding (NLU). Fine-tuning dataset is tokenized using GPT2 tokenizer. Each model variant is fine-tuned on downstream task. Fine-tuning is performed on datasets for $100$ epochs with batch size of $32$ or $256$ depending on dataset size, learning rate of $10^{-5}$ with FP16 data type. Pre-trained model weights are frozen and the last token pooling methodology is adopted by replacing the Language-Model (LM) head with a new linear layer followed by a Softmax layer as a classification head across GLUE tasks. Fine-tuning of the new linear layer is done by minimizing Cross-entropy loss with respect to target labels. All the models as described above - Base NanoGPT model, Symmetric-Attention model, NoiseFormer variants $V_1$ and $V_2$ - are fine-tuned on the GLUE dataset and evaluated on the downstream tasks described in the next section.

\section{Results}
In this section, we discuss the results of the pre-training and fine-tuning experiments of all the models described in the last section. We provide the results of our experiments and metrics on GLUE benchmark for all the models described above. Using any of the proposed NoiseFormer variants decreases the GPT2-based $NanoGPT$ weights by $\sim 5.77\%$ which is similar to Symmetric Attention model \cite{sym_attention}.

\subsection{Pre-training results - Analysis}
\begin{figure}[t]
     \begin{center}
          \centerline{\includegraphics[width=1.05\columnwidth]{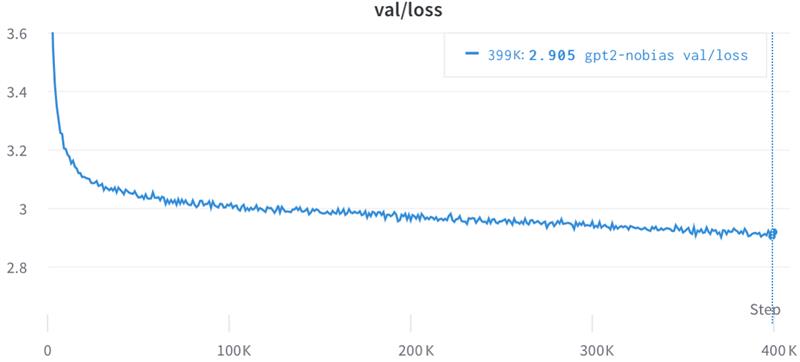}}
          \caption{Pre-training results for GPT2-based $NanoGPT$ model \cite{nanogpt2023}}
     \end{center}
     \vspace{-2em}
\end{figure}

Figure (8) depicts the results of pre-training on $NanoGPT$ base model \cite{nanogpt2023}. Figure (9) depicts the results of pre-training on the Symmetric Attention model \cite{sym_attention} and the proposed NoiseFormer variants $V_1$ and $V_2$ in terms of validation loss. It is observed that base model pre-trained for $300K$ steps achieved validation loss of $2.9$. In the case of Symmetric Attention model which is pre-trained for $100K$ steps, validation loss of $3.077$ is achieved. The proposed model variant $V_1$ achieves a loss of $3.115$ while $V_2$ achieves a loss of $3.069$ after $100K$ steps. From Figure (9), $V_2$ is performing better than Symmetric Attention model using different noise tensors across different heads within $MHA$ block, which is quite intuitive because every head masters in a particular aspect, while $V_1$ which models uniform noise across all heads produces greater validation loss. From Figure (8), it is to be noted that the validation loss for base model is also little bit higher than $3.0$ for $100K$ steps, which is in line with the results of the proposed model variants.

From Figures (8, 9), the base model's validation loss drops significantly in the initial steps and then drops with a lower rate later while the rate of decay of the loss is rather smooth for proposed model variants (of course, with an increased rate at the start). Over time, the proposed model variants achieve loss on par with the base model.

\begin{figure}[t]
     \begin{center}
          \centerline{\includegraphics[width=1.05\columnwidth]{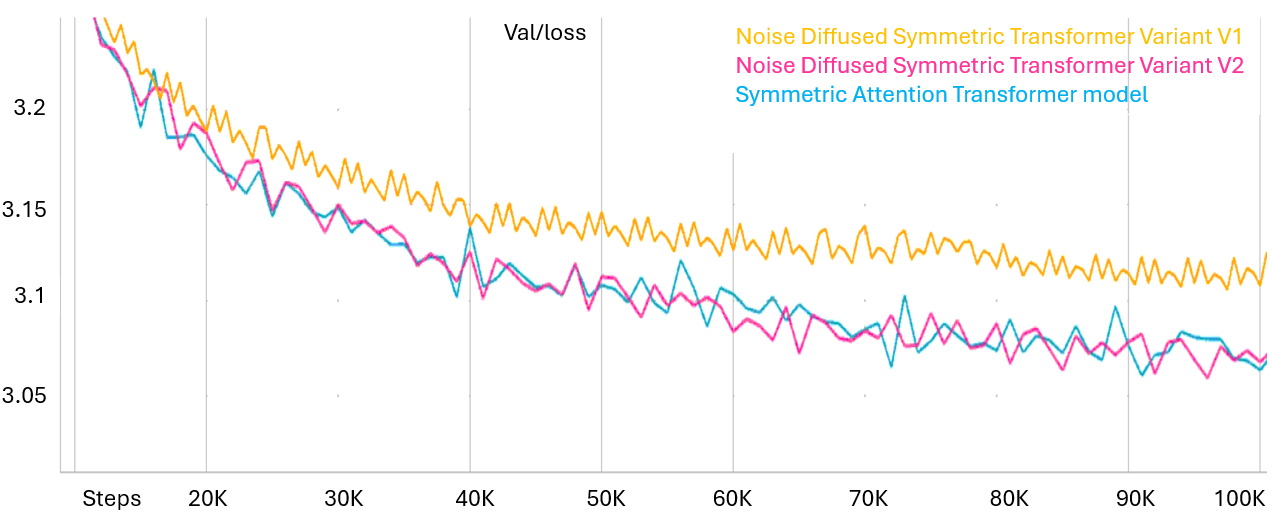}}
          \caption{Pre-training results for Symmetric Attention model \cite{sym_attention} and NoiseFormer variants}
     \end{center}
     \vspace{-2em}
\end{figure}

\subsection{Benchmarking with Fine-tuning - Analysis}
%Table 3 shows that the pair-wise variant performs better than the original variant with an increase
%of 0.6 points on the average GLUE score for both model sizes. The symmetric variant, however, is
%outperformed by the original variant in both cases, with a drop of 4 points on the average GLUE score.
%We also observe that both proposed variants have a lower standard deviation on the bert-base model.

Table II enumerates the performance results after fine-tuning the aforementioned models on the following GLUE benchmarking tasks. Each individual task has its own set of metrics and the performance on each of them is elaborated below:

\begin{itemize}
  \item $MRPC$ (Microsoft Research Paraphrase Corpus): This task is to decide whether two sentences are paraphrases or not. The metric being chosen as accuracy, from the results, the proposed NoiseFormer variants $V_1$ and $V_2$ perform better than Symmetric Attention model and lose to the Base model very narrowly by nearly $1-2\%$.  
  \item $MNLI$ (Multi-Genre Natural Language Inference): This task is based on 3-way classification between hypothesis-premise pairs based on Natural Language Inference. The results depict that the Base model performs better than proposed models and Symmetric Attention model while the proposed model variants get very closer to the Symmetric attention model in terms of accuracy.
  \item $RTE$ (Recognizing Textual Entailment): This task measures the binary entailment between sentence pairs with the proposed models closing the gap with Base model and performing better than Symmetric attention model in terms of accuracy, as shown in Table II.
  \item $QNLI$ (Question-answering Natural Language Inference): This task is based on 2-way classification between question-answer pairs respectively based on Natural Language Inference. From the results, it is seen that Symmetric Attention model performs better than all the other models and the proposed models get very closer to the Base model in terms of accuracy.
  %\item $COLA$ (Corpus of Linguistic Acceptability): The metric being considered as Mathew’s correlation, Symmetric Attention model performs far better than the base model and proposed model variants.
  %\item $QQP$ (Quora Question Pairs): The tasks are based on  question-question pair classification and sentiment analysis respectively, the proposed models beat the Base model while Symmetric attention model performs better than all the models in terms of accuracy.
  \item $SST2$ (Stanford Sentiment Treebank): This task is based on sentiment analysis. The proposed models beat the Base model while Symmetric attention model performs better than all the models in terms of accuracy as reflected in Table II.
\end{itemize}

The unweighted average GLUE score across all the above 5 benchmarking tasks is listed for the aforementioned models in Table II. From Table II, it is evident that the proposed NoiseFormer variants achieve higher GLUE scores as compared to Symmetric Attention model with respect to the Base model. %%sairam sairam sairam
The results convey that the proposed model takes a middle ground between the Base model and the Symmetric Attention model in terms of accuracy across tasks. The results suggest that the proposed model variants are trying to recover the loss in performance of Symmetric Attention model as compared to the Base model on tasks where Base model dominates Symmetric Attention model. For the tasks where Symmetric Attention model is performing better, the reduction in parametric dimensionality like $Q=K$ is playing a crucial role in extracting the right correlations between the parts of input prompts thus making the idea of Symmetric attention and the proposed NoiseFormers more attractive for many applications.

\begin{table}[htbp]
\caption{GLUE Benchmarks for downstream tasks with different models} %%sairam
\begin{center}
\begin{tabular}{|p{1.9cm}|p{0.63cm}|p{0.55cm}|p{0.5cm}|p{0.5cm}|p{0.5cm}|p{1.32cm}|}
\hline
\textbf{Model} & \textbf{MRPC} & \textbf{MNLI} & \textbf{RTE} & \textbf{QNLI} & \textbf{SST2} & \textbf{Avg GLUE} \\
\hline
$NanoGPT_{Base}$ & \textbf{74.75} & \textbf{56.91} & \textbf{61.01} & 67.30 & 81.65 & \textbf{68.324} \\
\hline
$SymAttn$ & 68.87 & 56.05 & 52.7 & \textbf{68.16} & \textbf{87.04} & 66.564 \\
\hline
$NoiseFormer_{V_1}$ & \textbf{73.28} & \textbf{55.74} & \textbf{58.12} & \textbf{66.88} & \textbf{82.68} & \textbf{67.34} \\
\hline
$NoiseFormer_{V_2}$ & 72.79 & 55.26 & 56.67 & 66.73 & 82.56 & 66.802 \\
\hline
\end{tabular}
\end{center}
\end{table}

%%sairam sairam sairam
We have also deployed the above mentioned pre-trained models for the task of inference-time text generation for a given input prompt. %- \textit{''What is the answer to life, the universe, and everything?''}. 
The maximum tokens for generation is set as $1000$. The inference is run on the same 8 Intel Gaudi2 distributed set-up as explained in the last section. The sequence length for the pre-trained model is $1024$. The temperature setting for inference is set as $0.8$ for random sampling of next token. We have generated $5$ inference samples, for the same input prompt across the models. We performed the experiment $5$ times and the execution times are averaged across the runs and across the sample generations. The memory and execution time of the models for inference-time text generation are also enumerated in the following Table (III). The reduction in compute can be attributed to the avoidance of $K$ matrix computation since $K=Q$ for Symmetric Attention and the proposed model variants.

\begin{table}[ht]
\caption{Inference Time Text generation - Performance} %%sairam
\begin{center}
\begin{tabular}{|p{3.0cm}|p{2.75cm}|p{2.0cm}|}
\hline
\textbf{Model} & \textbf{Model size} & \textbf{Wall-clock time for Inference} \\
\hline
$NanoGPT_{Base}$ & $123.5 \hspace{0.1cm} M$ & $15 \hspace{0.1cm} msec$ \\
\hline
$NanoGPT_{SymAttn}$ & $116.51 \hspace{0.1cm} M$ & $10 \hspace{0.1cm} msec$ \\
\hline
$NanoGPT_{Proposed (V_2)}$ & $116.51 \hspace{0.1cm} M$ & $10 \hspace{0.1cm} msec$ \\
\hline
\end{tabular}
\end{center}
\end{table}

\section{Further work} %%sairam sairam sairam
In this paper, we introduced Noise Diffused Symmetric Attention Transformer (NoiseFormer) which models generic attention mechanism as noisy symmetric attention. The proposed model builds upon the Symmetric Attention based Transformer with very minimal overhead in terms of memory and computational complexity, while enhancing the performance of the model. Going forward, we would like to empirically model the noise distribution involved between generic and symmetric attention across several training experiments with SOTA models thus achieving an optimal recipe for noise, in terms of mixture modes and densities, which gets diffused into the Attention matrices in the proposed architecture. As part of future work, we would like to see the impact of KL-divergence importance factor on the performance in terms of diversity of generated samples and accuracy. Also, we would like to extend the proposed model with input-conditioned noise models to analyze the impact on accuracy with computational trade-off.

%Further, we would like to analyze the generic attention as noisy attention empirically which enables more accurately modelling the noise distribution which paves the way for optimal noise distribution to be diffused into the Attention matrices in the proposed architecture.

% \vspace{100pt}

%%sairam sairam sairam
\section{Acknowledgment}
The authors would like to thank Aditya, Rathore from IIIT, Hyderabad for developing the fine-tuning scripts and assisting with the creation of the paper's figures.


\begin{thebibliography}{1}

     \bibitem{gqa2023}
     J. Ainslie, J. Lee-Thorp, M. de Jong, Y. Zemlyanskiy, F. Lebron, S. Sanghai, ``GQA: Training Generalized Multi-Query Transformer Models from Multi-Head Checkpoints,''
     \emph{Proceedings of the 2023 Conference on Empirical Methods in Natural Language Processing}

     \bibitem{mqa2021}
     N. Shazeer, ``Fast Transformer Decoding: One Write-Head is All You Need,''
     \emph{arXiv:1911.02150}, 2019.

     \bibitem{longformer2020}
     I. Beltagy, M. E. Peters, A. Cohan, ``Longformer: The Long-Document Transformer,''
     \emph{arXiv:2004.05150v2}, 2020.

     \bibitem{reformer2020}
     N. Kitaev, L. Kaiser, A. Levskaya, ``Reformer: The Efficient Transformer,''
     \emph{ICLR}, 2020.

     \bibitem{deepseek_mla2024}
     DeepSeek-AI, ``DeepSeek-V2: A Strong, Economical, and Efficient Mixture-of-Experts Language Model,''
     \emph{arXiv:2405.04434v5}, 2024.

    \bibitem{sym_attention}    
    M. Courtois, et al., ``Symmetric Dot-Product Attention for Efficient Training of BERT Language Models'', 
    \emph{Findings of ACL 2024}
  
     \bibitem{attention}
     A. Vaswani, et al., ``Attention Is All You Need,''
     \emph{Advances in neural information processing systems, 30}, 2017.
     
     \bibitem{DDPM}
     J. Ho, A. Jain, P. Abbeel, ``Denoising Diffusion Probabilistic Models''
     \emph{NeurIPS 2020}.
     
     \bibitem{nanogpt2023}
     A. Karpathy, ``nanoGPT: The simplest, fastest repository for training/finetuning medium-sized GPTs,''
     \emph{GitHub repository}, 2023. [Online]. Available: {https://github.com/karpathy/nanoGPT}.

     \bibitem{vae2022}
     D. P. Kingma, M. Welling, ``Auto-Encoding Variational Bayes,''
     \emph{ICLR 2014}.

     \bibitem{free}
     F. Fleuret, ``The Free Transformer,''
     \emph{ICLR 2026}

     %\emph{}, 2023

\end{thebibliography}
\end{document}